\documentclass[10pt, a4paper]{article}
\usepackage{multirow} 
\usepackage{makecell}
\usepackage{float}
\usepackage{lrec-coling2024} 

\usepackage{tablefootnote}
\usepackage{caption}
\title{Probing Large Language Models for Scalar Adjective Lexical Semantics and Scalar Diversity Pragmatics}

\name{Fangru Lin, Daniel Altshuler, Janet B. Pierrehumbert} 

\address{University of Oxford\\
         Oxford, United Kingdom \\
         \{fangru.lin, daniel.altshuler\}@ling-phil.ox.ac.uk, janet.pierrehumbert@oerc.ox.ac.uk}

\abstract{
Scalar adjectives pertain to various domain scales and vary in intensity within each scale (e.g. \emph{certain} is more intense than \emph{likely} on the likelihood scale). 
Scalar implicatures arise from the consideration of alternative statements which could have been made. They can be triggered 
by scalar adjectives and require listeners to reason pragmatically about them. Some scalar adjectives are more likely to trigger scalar implicatures than others. This phenomenon is referred to as scalar diversity. In this study, we probe different families of Large Language Models such as GPT-4 for their knowledge of the lexical semantics of scalar adjectives and one specific aspect of their pragmatics, namely scalar diversity. We find that they encode rich lexical-semantic information about scalar adjectives. However, the rich lexical-semantic knowledge does not entail a good understanding of scalar diversity. We also compare current models of different sizes and complexities and find that larger models are not always better. Finally, we explain our probing results by leveraging linguistic intuitions and model training objectives.
 \\ \newline  }

\begin{document}

\maketitleabstract
\section{Introduction}

Scalar adjectives (SAs) are words such as \emph{likely}, \emph{certain}, \emph{warm}, and \emph{scalding}. They describe different scales of properties. For instance, \emph{warm} and \emph{scalding} describe temperature, while \emph{likely} and \emph{certain} describe probabilities. Scalar adjectives can describe the same scale while differing in intensity. For example, \emph{certain} is more intense than \emph{likely} on the likelihood scale because it is used to make a logically stronger statement about a given situation. 

\begin{figure}[!h]
    \centering
    \includegraphics[scale=0.25]{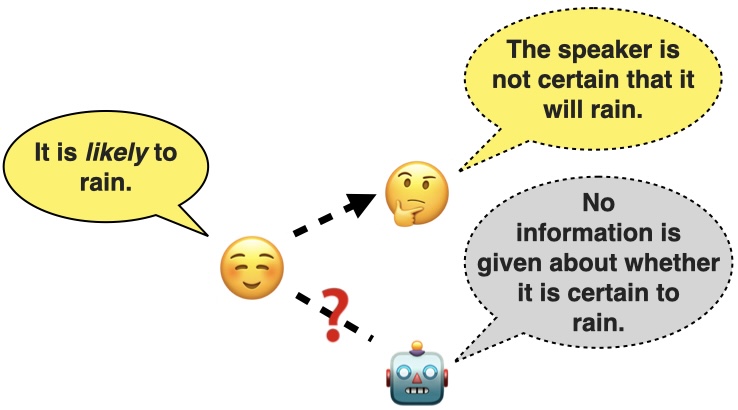}
    \caption{People often mean more than what they literally say. Humans can easily infer implied messages, while LLMs often fail to do so.}
    \label{fig:task-illustration}
\end{figure}

Scalar implicatures (SIs) arise from the
consideration of alternative statements that could have been made (Figure~\ref{fig:task-illustration}). They can be triggered by SAs. For instance, when a speaker utters ‘It is \emph{likely} to rain’, a hearer may conclude that the speaker is \emph{not certain} that it will rain. In particular, the hearer may reason that the speaker could have provided the logically stronger statement, ‘it is \emph{certain} to rain’, but did not do so. Psycholinguistic studies indicate that some SAs are more likely to generate implicatures than others \cite{vt2016, ng, ronai2022three}. This phenomenon is referred to as \emph{scalar diversity}. For instance, \emph{likely} tends to indicate \emph{not certain}, while \emph{good} does not tend to indicate \emph{not excellent}.

SI is a long-standing topic of research in pragmatics because it reveals fundamental aspects of human linguistic and cognitive capabilities in areas such as the Theory of Mind \cite{feng2021understanding}. Accordingly, SIs pose important challenges for the development of NLP models with human-like capabilities \cite{sap-etal-2022-neural}. SIs are also important in practice for downstream tasks that include Natural Language Inference \cite{williams-etal-2018-broad}, Sentiment Analysis \cite{socher2013recursive}, and indirect Question Answering \cite{de-marneffe-etal-2010-good}. State-of-the-art large Language Models (LLM) such as GPT-4\footnote{\url{https://openai.com/research/gpt-4}} have remarkable performance on many classic benchmarks. However, they have proved to be fragile in some semantic and pragmatic tasks that are easy for humans \cite{liu2023we, lin2024graph}. The goal of this paper is to probe this discrepancy with an in-depth investigation of SIs with SAs. 

Recent research on LLMs' understanding of SA has focused on: (i) probing the lexical semantics of SAs  \cite{Liu_Xiang_Ding_2023} or (ii) their scalar diversity \cite{hu-etal-2022-predicting,10.1162/tacl_a_00579}. In this paper, we investigate both (i) and (ii). We first utilize two probing methods to evaluate LLMs' knowledge about SAs' scale membership and intensity information. Using previously published datasets, we then assess whether LLMs show human-like scalar diversity in judgments about SA items. Finally, we compare LLMs' pragmatic and lexical-semantic knowledge and explain our observations based on linguistic theory.

Our paper includes three main findings. First, LLMs generally encode rich lexical-semantic information about SAs (\S \ref{sec:lexical-semantics}). Second, LLMs have unsatisfying performance in capturing scalar diversity despite encoding rich lexical-semantic information about SAS (\S \ref{sec:scalar-diversity}). Third, the size of the LLMs does not correlate with how well they perform on our tasks (\S \ref{sec:lexical-semantics} and \S \ref{sec:scalar-diversity}): 
While the increase in model size is sometimes claimed to invariably improve performance (the so-called ''scaling law"), in our study, some large models do worse than smaller models that have different architectures or training objectives.\footnote{Code is in \url{https://github.com/fangru-lin/llm_scalar_adj}.}

\section{Related works}
\subsection{Scalar Adjective Lexical Semantics}

Since the seminal work by \citet{kamp}, SAs have received a lot of attention in formal semantics and the philosophy of language. The same cannot be said about NLP research, where SAs have only recently garnered attention. 
Some of this research has been concerned with the lexical semantics of SAs, particularly focusing on their scale membership and their intensities \cite{demelo, kim2013deriving, shivade2015corpus, wilkinson, cocos-etal-2018-learning}. One feature of this research is the use of static rather than contextualized word embeddings from LLMs. This is important because \citet{gari-soler-apidianaki-2020-bert} have found that BERT-base \cite{devlin-etal-2019-bert} contextualized word embeddings encode richer information about scalar intensities in vector space. Nevertheless, \citet{Liu_Xiang_Ding_2023} report that many state-of-the-art models, even after fine-tuning on MNLI \cite{williams-etal-2018-broad}, have unsatisfying adjective degree estimations in textual inference.

\subsection{Scalar Implicature and Scalar Diversity Pragmatics}
In a study of SIs, \citet{schuster-etal-2020-harnessing} have found that LSTM \cite{hochreiter1997long} can be trained to infer \textit{some-not all} implicatures (e.g. ‘Some student cheated’ implicates that not all students cheated). \citet{jeretic-etal-2020-natural} report that BERT fine-tuned on MNLI nearly always predicts that the determiner \emph{some} entails \emph{not all},  but that SAs are treated as synonyms regardless of their intensity. This means that the model may lack relevant pragmatic knowledge about adjectives. In contrast, two recent studies \cite{hu-etal-2022-predicting, 10.1162/tacl_a_00579} achieved more favorable results with surprisal measures derived from GPT2. String-based surprisal, which considers the surprise level of a given strong word to appear as an alternative to a weak word out of all possible strings in a context, is estimated by the likelihood of a strong word to appear in its position by a GPT-2 model \cite{radford2019language}. Concept-based surprisal treats an alternative as a member of a string set with similar concepts. The surprisal rate for the alternative is considered as an average over that of all strings in such a set. Using these two measures, they found that scalar diversity correlates with string-based and concept-based surprisal.   \citet{ruis2024goldilocks} tested different LLMs on general conversational implicature understanding. They found that GPT-4 is the best-performing model with 30-shot chain-of-thought prompting \cite{wei2022chain}, achieving 88.66\% in generalized implicature calculation. This is the category that our adjective-triggered SIs fall under.

\subsection{Direct and Indirect Probing}
Direct and indirect probing methodologies are widely used to understand LLMs' knowledge. Direct probing is used to analyze the hidden representations encoded in LLMs \cite{gari-soler-apidianaki-2020-bert}. This is only possible when the LLM is open-source. Indirect probing is used to analyze the representations of closed-source models. This involves testing performance on various semantic tasks. Typical tasks use textual prompts to assess the prompted answers \cite{ettinger-2020-bert}. \citet{petroni-etal-2019-language} argue that indirect probing only reveals the lower bound of model capabilities.  Follow-up works have provided various methods to increase the bound \cite{zhong-etal-2021-factual, white2023prompt}.

We build on these results as follows. Since direct probing provides a good estimate of a model's competence, but closed-source LLMs only allow for indirect probing,  we use both methods for some open-source models. We then use the results on these models to ground the results of indirect probing for other models.  We make the following assumption: if the two probing methods show similar trends regarding model performance for the open-source models, we deem our indirect probing methods as valid in comparing the relative capabilities of different closed-source models.

\section{Probing Lexical Semantics}
\label{sec:lexical-semantics}
Recent research on the lexical semantics of SAs have mostly focused on deriving their intensity information from open-source models \cite{gari-soler-apidianaki-2020-bert, Liu_Xiang_Ding_2023}. In this section, we describe how we probe three model families, 
for two aspects of SA lexical semantics: (i) scale membership (e.g. \emph{likely} and \emph{certain} are on the likelihood scale), and (ii) adjective intensity (e.g. \emph{certain} is more intense than \emph{likely} within the same scale).

We assess three model families, containing eight models in total, to understand how different model architectures, training objectives, and sizes affect lexical-semantic knowledge. All models except GPT-4 are accessed via huggingface library \cite{wolf-etal-2020-transformers}, and GPT-4 is accessed via OpenAI API.\footnote{See appendix \ref{sec:implementation_detail} for more implementation details.}

\textbf{Encoder Models} BERT-base/large(b/l) (110M/340M) \cite{devlin-etal-2019-bert}, RoBERTa-base/large(b/l) (123M/354M) \cite{liu2019roberta}

\textbf{Decoder Models} Falcon-7B-instruct (Falcon) \cite{falcon}, GPT-4\footnote{We do not use LLaMA \cite{touvron2023llama} due to our institutional requirement. We use gpt-4-0613 for GPT-4.}

\textbf{Encoder-decoder Models} Flan-T5-xl/xxl (3.5B/11B) \cite{flant5}

\subsection{Datasets for Lexical-Semantic Probing}
\subsubsection{Scalar Adjective Datasets} 

Following \citet{gari-soler-apidianaki-2020-bert}, we use three SA datasets: DEMELO (\textbf{DM}) \citelanguageresource{DEMELO}, CROWD (\textbf{CD}) \citelanguageresource{CROWD}, and WILKINSON (\textbf{WK}) \citelanguageresource{WILKINSON} which contain 185 half-scales in total (Table~\ref{tab:dataset}). For example, the positive side of the quality scale ranges from \emph{good} to \emph{awesome}. Each contains different categories of SAs. Some contain adjectives describing physical appearance while others do not.

\begin{table}[ht]
    \centering
    \begin{tabular}{c|c|c|c}
        &\textbf{DM}&\textbf{CD}&\textbf{WK} \\
        \hline
        \textbf{Half-scales}&87&77&21\\
        \hline
        \textbf{Distinct adjective pairs}&548&330&61\\
        
    \end{tabular}
    \caption{Overview of half-scale counts and distinct adjective pairs in SA datasets.}
    \label{tab:dataset}
\end{table}

\textbf{DM} Half-scales are first collected from WordNet 3.0 \citelanguageresource{WordNet} and then annotated for intensity by two native English speakers.

\textbf{CD} Adjectives are first collected from the Paraphrase Data Base \citelanguageresource{ganitkevitch-etal-2013-ppdb, PPDB} and then annotated by crowd workers. Annotators are asked to identify whether given adjectives are on the same scale for multiple rounds, and then annotate the intensity of adjectives.

\textbf{WK} Adjectives are collected via crowdsourcing. Crowd workers are first presented with prompt words that belong to a full scale and asked to list other semantically related adjectives. Scales are, then, cleaned automatically based on workers’ competence and annotated manually for adjective intensities. This study uses partitioned \textbf{WK} with half scales in \citet{cocos-etal-2018-learning}, which are derived from the Paraphrase Data Base \citelanguageresource{ganitkevitch-etal-2013-ppdb, PPDB}.

\subsubsection{Context Sentence Datasets} To compute contextualized word embeddings for adjectives, we use the context sentence dataset \textbf{ukWac} from \citetlanguageresource{ukWac}. \textbf{ukWac} provides 10 context sentences for each half-scale \emph{s} in \textbf{DM}, \textbf{CD}, and \textbf{WK}. All adjectives on \emph{s} share the same context set of 10 different sentences. Each candidate sentence contains an adjective on \emph{s} and allows lexical substitution of its scale-mates. The acceptability after substitution is ensured by computing their acceptability scores using context2vec \cite{melamud-etal-2016-context2vec}. An example is given below, where the italic word is an adjective on \emph{s}, and appears in the original context sentence. Other scale alternatives to it are inside the brackets.
\begin{quote}
    \texttt{They are extremely catchy and are not only great to listen to, but they are also \emph{thrilling} (interesting/moving/exciting) to sing.}
\end{quote}

\subsection{Probing Scale Membership}
In this subsection, we assess whether LLMs encode the notion of scale membership. Ideally, we would expect models to be able to assign adjectives to their corresponding scales in context (e.g. 
\emph{warm} belongs to the temperature scale).

\subsubsection{Scale Membership Direct Probing Method}
We directly probe scale membership by getting contextualized representations of SAs and half-scales and calculating their similarities. We retrieve word embeddings using \textbf{ukWac} by sampling different sentences for each adjective $a$ on a half-scale $s$ to obtain its contextualized representation $\vec{a}$. We derive a scale vector $\vec{s}$ for each $s$ by adding the representations of the weakest and strongest adjectives on $s$. For instance, for the scale \emph{adequate-fine-fitting-good}, where \emph{adequate} is the least intense and \emph{good} is the most intense adjective, $\vec{s}$ is $\vec{adequate}+\vec{good}$. Then, the cosine similarity $cos(\vec{a},\vec{s})$ between each adjective $a$ in a SA dataset $D_a$ and all other  $\vec{s}$ in the dataset is computed. The scales are ranked by $cos(\vec{a},\vec{s})$ for each adjective; the scale with the highest $cos(\vec{a},\vec{s})$ is considered to be the most likely scale for $a$ to belongs to, and so forth. For evaluation, the ranking of the scale $s$ that $a$ belongs to ($rank_s$) is considered. We consider Mean Reciprocal Rank (MRR) as the evaluation metric; see Equation (\ref{eq:rank}).

\begin{equation}\label{eq:rank}
    MRR = \frac{1}{|D_a|}\sum_{s\in{D_a}}\frac{1}{rank_s}
\end{equation}

Intuitively, the higher the scale that an adjective belongs to is ranked, the closer MRR will be to 1. For instance, if all adjectives are aligned on their corresponding scales, the MRR will be 1. If models fail to rank the correct scales as closest, it means that they do not encode word polysemes and fine-grained distinction among different scales.

\subsubsection{Scale Membership Direct Probing Experiment and Results}
When target adjectives are tokenized as multiple segments, token representations are averaged as the adjectives' final representations. We use layer-wise adjective word embeddings to evaluate information in each layer (e.g. $\vec{a}$ and $\vec{s}$ have 12 representations using a 12-layer model). The experiment is repeated ten times with different random seeds for context sentences. We use fast-text static word embeddings\footnote{\url{https://github.com/facebookresearch/fastText/}} \cite{mikolov2018advances} trained on 600B Common Crawl data as a baseline. For the baseline, we add the representations of strongest and weakest adjectives on \emph{s} as $\vec{s}$, then also rank $cos(\vec{a},\vec{s})$ for each \emph{a} to calculate MRR. Here, we only report the results for the best-performing layer for non-baseline models in Table \ref{tab:membership}.\footnote{We also tried other alternative methods to compute membership, which does not generally work as well as the method we report here. See Appendix~\ref{sec:alternative_membership}.}

\begin{table}[ht]
    \centering
\resizebox{\linewidth}{!}{%
    \begin{tabular}{c|c|c|c}
         \textbf{Models}& \textbf{DM} & \textbf{CD} & \textbf{WK} \\
         \hline
         \textbf{fast-text}& 0.842& 0.716& 0.983\\
         \hline
         \textbf{BERT-b} &0.829$_{\pm0.010}$&0.797$_{\pm0.010}$&\textbf{0.997$_{\pm0.004}$} \\
         \hline
         \textbf{BERT-l} &\textbf{0.853$_{\pm0.007}$}&\textbf{0.805$_{\pm0.011}$}&\textbf{0.997$_{\pm0.006}$}\\
         \hline
         \textbf{RoBERTa-b}&0.668$_{\pm0.014}$&0.705$_{\pm0.007}$&0.906$_{\pm0.018}$\\
         \hline
         \textbf{RoBERTa-l}&0.777$_{\pm0.011}$&0.757$_{\pm0.008}$&0.977$_{\pm0.010}$\\
    \end{tabular}}
    \caption{Direct scale membership probing results, subscripted numbers are standard deviation in ten runs. The best results per dataset are in bold.}
    \label{tab:membership}
\end{table}

The fast-text baseline is quite strong considering its training data size. All LLMs encode rich scale information as they mostly outperform (with some behaving on par with) the baseline. Model-wise, BERT models encode more information than RoBERTa. Within the same architecture, the bigger a model is, the better they are. Across datasets, \textbf{WK} is relatively easy because it (i) has fewer polysemous adjectives with the same forms on different scales and (ii) has fewer scales for classification than the other two datasets. 

\subsubsection{Scale Membership Indirect Probing Method}
Inspired by \citet{notwacky} and \citet{demelo}, we use four templates including \emph{ADJ$_{weak}$, if not ADJ$_{strong}$} to indirectly assess LLMs' knowledge about scale membership. Adjectives on the same scale are more likely to co-occur in these constructions than those that are not (e.g. \emph{warm, if not hot} should be more likely to appear than \emph{warm, if not thin}).\footnote{See Appendix \ref{sec:membership_template} for all templates.}

For each adjective on a half scale that is not the strongest item, we use it as \emph{ADJ$_{weak}$} in our templates and obtain 5 most likely words as \emph{ADJ$_{strong}$}. For instance, for scale
\emph{adequate-fine-fitting-good}, we prompt models to answer what is likely to be \emph{ADJ$_{strong}$} in \emph{adequate/fine/fitting, if not \emph{ADJ$_{strong}$}}. We consider a case to be correct if any top-5 word in \emph{ADJ$_{strong}$} is on the same scale with \emph{ADJ$_{weak}$}, and the overlap is not trivial (e.g. \emph{adequate or even adequate} is a trivial completion, \emph{good or even adequate} is considered to be correct despite the wrong intensity ranking).

\subsubsection{Scale Membership Indirect Probing Experiment and Results} For encoder models, we put [MASK] and a comma in the position of  \emph{ADJ$_{strong}$} (e.g. \emph{ADJ$_{weak}$, if not [MASK],}) as preliminary experiment results show that BERT mostly predicts punctuation marks as top 5 predictions for [MASK]  when it appears as the last token. The comma is used because it does not constrain whether its preceding adjective is attributive (i.e. \emph{can} be immediately followed by a noun) or predicative (i.e. \emph{cannot} be immediately followed by a noun). For non-baseline models except GPT-4, we directly generate \emph{ADJ$_{strong}$} given the preceding prompt. We use a similar generation objective for GPT-4 with the best template in other models (see appendix \ref{sec:gpt4_prompt}). We use the Google Ngram corpus slice for 2019 \cite{lin-etal-2012-syntactic} as the indirect probing baseline with a similar setting, to obtain the 5 most likely completions for \emph{ADJ$_{strong}$} in each construction. In Table \ref{tab:indirect_probe_res}, we report the best results among all templates in the assessed dataset (we report results based on selecting the best templates with held-out datasets in Table~\ref{tab:indirect_membership_app}).
\begin{table}[h]
    \centering
    \begin{tabular}{c|c|c|c}
         \textbf{Models} & \textbf{DM} & \textbf{CD} & \textbf{WK} \\
         \hline
         \textbf{Google Ngram}&0.287$_{1}$&0.075$_{1}$&0.368$_{1}$\\
         \hline
         \textbf{BERT-b}&0.425$_{1}$&0.066$_{1}$&0.167$_{1}$\\
         \hline
         \textbf{BERT-l}&0.504$_{1}$&0.109$_{3}$&0.250$_{1}$\\
         \hline
         \textbf{RoBERTa-b}&0.271$_{4}$&0.064$_{3}$&0.114$_{1}$\\
         \hline
         \textbf{RoBERTa-l}&0.463$_{1}$&0.100$_{1}$&0.286$_{1}$\\
         \hline
         \textbf{Falcon}&0.265$_{1}$&0.045$_{1}$&0.167$_{1}$\\
         \hline
         \textbf{GPT-4}&0.540&\textbf{0.273}&\textbf{0.500}\\
         \hline
         \textbf{Flan-T5-xl}&\textbf{0.544$_{3}$}&0.177$_{3}$&\textbf{0.500$_{3}$}\\
         \hline
         \textbf{Flan-T5-xxl}&0.515$_{3}$&0.156$_{3}$&0.364$_{3}$\\
    \end{tabular}
    \caption{Indirect probing results for scale membership. The best results per dataset are in bold. Subscripts refer to the indices of the best templates.}
    \label{tab:indirect_probe_res}
\end{table}

Indirect probing always underperforms direct probing for all models. This result aligns with \citet{petroni-etal-2019-language}. The relative rankings of different models are largely consistent. BERT is better than RoBERTa given the same size class, and larger encoder models are better than smaller ones with the same architecture. These observations support the validity of our methods.

We find that all models except Falcon encode rich information about scale membership, in that they either behave on par with the baseline or outperform it. GPT-4 and Flan-T5-xl each achieve the best results on two benchmarks.  While GPT-4 has competitive results with Flan-T5-xl in \textbf{DM} and \textbf{WK}, it is much better for \textbf{CD}. Therefore, we conclude that GPT-4 is the best-performing model in this task. Within the same model family, the scaling law tends to hold (i.e., larger is better). However, scaling does not fully explain the pattern of results. For instance, Falcon underperforms much smaller encoder models and even the baseline, and Flan-T5-xxl underperforms Flan-T5-xl.

\subsection{Probing Scalar Intensity}
\label{sec:scalar-intensity}
In this section, we describe how we probe whether LLMs understand that SAs on the same scale have varied intensities (e.g. \textit{hot} denotes a higher temperature than \textit{warm}).

\subsubsection{Scalar Intensity Direct Probing Method}
\begin{figure*}[ht]
    \centering
    \includegraphics[scale=0.5]{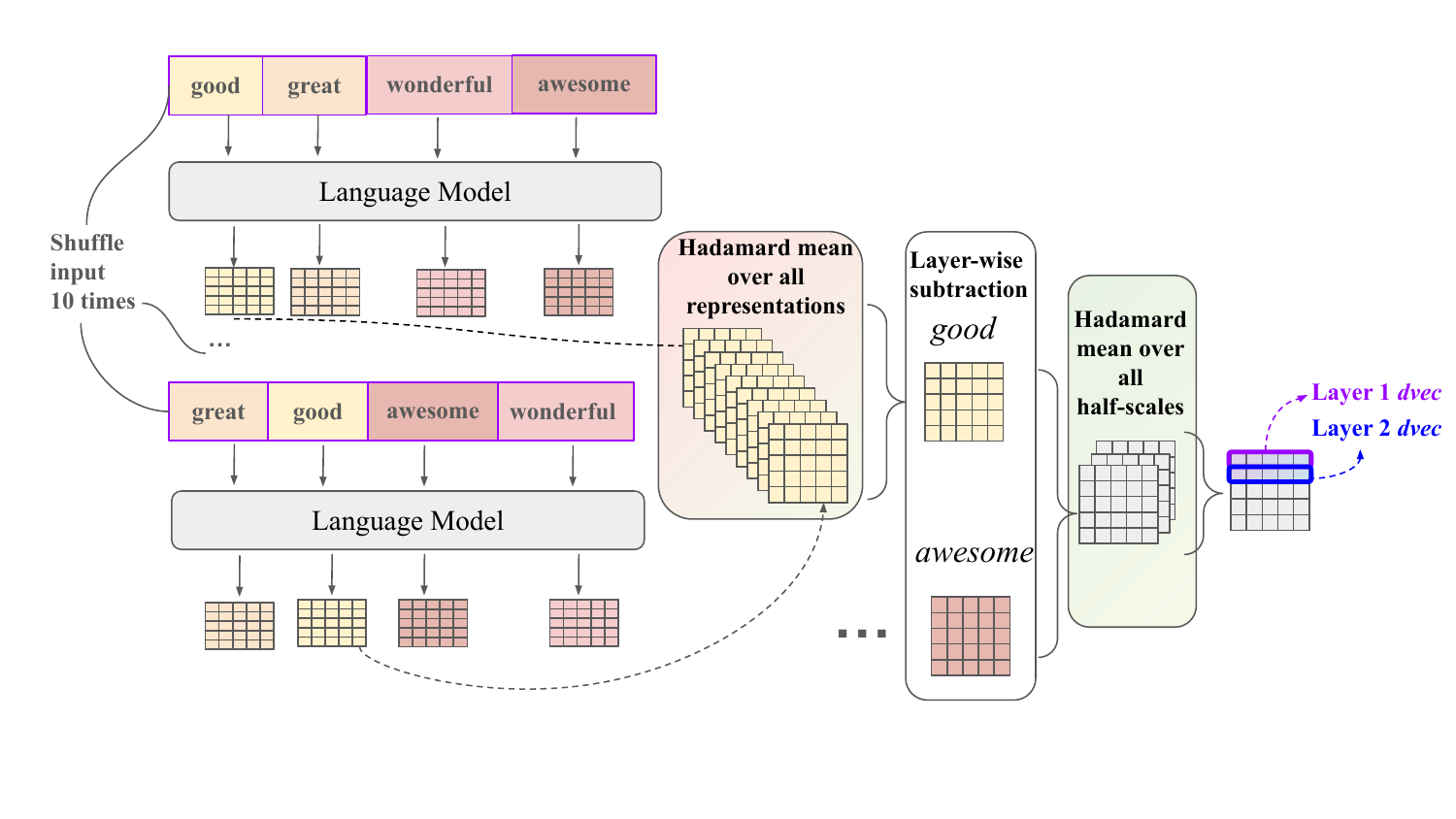}
    \caption{The process to derive intensity vector $\vec{d_{vec}}$. First, an adjective half-scale is randomly shuffled ten times for the order of adjectives as inputs to a language model. Then the encoded word vectors for the same word in different inputs are conducted with the Hadamard mean to derive the final representation of the word. After that, intensity vector $\vec{d_{vec}}$ is calculated by subtracting layer-wise representation of the weakest adjective from the strongest adjective ($\vec{awesome}-\vec{good}$ in this case) then averaging over all relevant half-scale subtractions in a dataset. Then layer-wise $\vec{d_{vec}}$ is used to probe language models' knowledge for adjective intensities.}
    \label{fig:intensity-vector}
\end{figure*}
Unlike previous methods which ground SAs in identical contexts and calculate them in separate runs (e.g. It is a \emph{good/great/wonderful/awesome} movie) \cite{gari-soler-apidianaki-2020-bert} (G\&A), we use a novel method, which involves binding all SAs on the same scale as one single input to obtain their contextualized representations (e.g. \emph{good great wonderful awesome}), as illustrated in Figure \ref{fig:intensity-vector}. Our assumption is that people are more likely to notice scalar words have different intensities when they are salient in the same context \cite{degreeestimateronaixiang}. 

For each half-scale $s$, we shuffle and bind all SAs $a$ on $s$ ten times to get 10 different inputs (e.g. \emph{good great wonderful awesome}, \emph{great wonderful good awesome}, etc.). For each $a$, we average the representations of it in 10 inputs to get its final representation. This is to avoid potential heuristic associations between word order and scalar intensity.

After getting the representation of all $a$, we utilize the best method (DiffVec) in \citet{gari-soler-apidianaki-2020-bert} (G\&A) to get the intensity ranking for them. Specifically, the intensity of all adjectives $a$ in dataset $D_a$ is measured by their cosine similarities to an external global intensity vector $\vec{d_{Vec}}$ derived from another SA dataset $D_{vec}$ ($D_{vec} \neq D_{a}$).\footnote{We don't use the global intensity vector in $D_{a}$ due to the concern of generalisability.}

For every half-scale $s$ in $D_{vec}$, the mildest and the extreme adjectives on $s$ are notated as $a_{ms}$ and $a_{es}$. We calculate $\vec{d_{Vec}}$ with Equation (\ref{eq:dvec}).

\begin{equation}
    \vec{dVec} = \frac{1}{|D_{vec}\setminus{D_{a}}|}\sum_{s\in{D_{vec}\setminus{D_{a}}}}{{\vec{a_{es}}}-{\vec{a_{ms}}}}
    \label{eq:dvec}
\end{equation}

For all $s \in D_a$, adjectives on $s$ are ranked by their cosine similarities to $\vec{d_{Vec}}$: higher similarity means higher intensity. If two adjectives have the same similarity, they are treated as equally intense. 

\subsubsection{Scalar Intensity Direct Probing Experiment and Results} 

We probe layer-wise word embeddings in BERT-b/l, RoBERTa-b/l and report the best results for pair-wise accuracy across layers using $\vec{d_{Vec}}$ in different datasets. We again provide fast-text embeddings as a baseline here. For the baseline, we use static embeddings of $\vec{a_{es}}$ and $\vec{a_{ms}}$ for all $s\in{D_{vec}\setminus{D_{a}}}$ to derive $\vec{dVec}$ as in Equation \ref{eq:dvec}.

Results are averaged over adjective pairs and shown in Table \ref{tab:direct_intensity}.

\begin{table}[ht]
    \centering
    \resizebox{\linewidth}{!}{%
    \begin{tabular}{c|c|c|c|c}
         \textbf{Model}&\textbf{Method}&\textbf{DM}&\textbf{CD}&\textbf{WK}  \\
         \hline
         \textbf{fast-text}&-&0.637$_{WK}$&0.685$_{DM}$&0.836$_{CD}$\\
         \hline
         \multirow{2}{*}{\textbf{BERT-b}}&G\&A&\underline{0.646$_{CD}$}&\underline{0.735$_{DM}$}&0.902$_{DM}$\\
         &Ours&0.639$_{CD}$&0.706$_{DM}$&\textbf{\underline{0.967$_{DM}$}}\\\hline
         \multirow{2}{*}{\textbf{BERT-l}}&G\&A&\textbf{\underline{0.695$_{CD}$}}&\underline{0.731$_{DM}$}&\underline{0.918$_{DM}$}\\
         &Ours&0.673$_{CD}$&0.727$_{DM}$&0.902$_{DM}$\\\hline
         \multirow{2}{*}{\textbf{RoBERTa-b}}&G\&A&0.557$_{WK}$&0.645$_{DM}$&0.820$_{DM}$\\
         &Ours&\underline{0.648$_{CD}$}&\underline{0.748$_{DM}$}&\underline{0.934$_{DM}$}\\\hline
         \multirow{2}{*}{\textbf{RoBERTa-l}}&G\&A&0.595$_{CD}$&0.682$_{DM}$&0.836$_{DM}$\\
         &Ours&\underline{0.664$_{CD}$}&\textbf{\underline{0.752$_{DM}$}}&\underline{0.934$_{DM}$}\\
    \end{tabular}}
    \caption{Direct scalar intensity probing with our method and G\&A on different models. Subscripted letters refer to the source dataset of $\vec{d_{Vec}}$. The best results per dataset are in bold. The best results using the same model are underlined.}
    \label{tab:direct_intensity}
\end{table}

We find that our simple method is generally better than G\&A.\footnote{Our method is also superior to \citet{gari-soler-apidianaki-2020-bert} in other metrics. See appendix \ref{sec:intensity_ranking_full_res}.} Globally, our method achieves the best-ranking results in two out of three datasets. Although G\&A marginally outperforms our method in some of the datasets in BERT, our method uniformly largely outperforms G\&A in RoBERTa, showing that it is more robust across different models and datasets.

All models encode rich information about SA intensity by outperforming the baseline. Across models, BERT models again dominate in two out of three datasets, showing that they encode more intensity information than RoBERTa, which indicates that larger models are not always better. Across datasets, \textbf{WK} is again relatively easy because it does not contain ties.

\subsubsection{Scalar Intensity Indirect Probing Method}
To complement closed-source models, we indirectly probe intensity using a similar objective as in \citet{notwacky}. Specifically, for each pair of SAs to compare intensities, we compare the perplexity of minimal-pair prompts containing these adjectives such as '\emph{good} but not \emph{awesome}' and '\emph{awesome} but not \emph{good}', with the former featuring correct order of scalar adjectives and the latter incorrect. Ideally, models should always compute lower perplexity for the first prompt, as the latter is infelicitous or even ungrammatical.

We adopt all 34 templates used in the data collection of \citet{demelo}. \footnote{Except \emph{ADJ$_{strong}$(,) or very ADJ$_{weak}$} because some adjectives examined in this study are extreme ones \cite{paradis1998degree} which cannot be modified by \emph{very} (e.g. \emph{very terrific}). See Appendix \ref{sec:intensity_template} for the full template list.} For each template, we compare the perplexity of correct and incorrect constructions. We consider that LLMs encode the scalar intensity for the corresponding adjective pair if the correct one has lower perplexity. We consider two adjectives to be equally intense if switching their positions in a template produces equally unlikely phrases.

\subsubsection{Scalar Intensity Indirect Probing Experiment and Results} 

For encoder models, we estimate pseudo-perplexity by masking out tokens one by one and multiplying the likelihood of the original token appearing in the mask position to compute the sequence's pseudo-likelihood \cite{salazar-etal-2020-masked}. For other models except GPT-4, we directly compute perplexity for each sequence. For GPT-4, we use the best-performing template for other models and mimic the setting as in Appendix \ref{sec:gpt4_prompt}. As a baseline, we use Google Ngram corpus slice in 2019 to retrieve the likelihood of constructions as an analogy to perplexity. We then consider a case to be classified correctly if the correct construction is more likely to appear than the incorrect one.  We use pair-wise accuracy as an evaluation metric and report the best results among all templates in the assessed dataset. Final results are averaged over pairs and shown in Table \ref{tab:indirect_intensity} (we report results based on selecting the best templates with held-out datasets in Table~\ref{tab:indirect_intensity_app}).

\begin{table}[ht]
    \centering
\resizebox{\linewidth}{!}{%
    \begin{tabular}{c|c|c|c}
         \textbf{Models} & \textbf{DM} & \textbf{CD} & \textbf{WK}\\
         \hline
         \textbf{Google ngram}&0.312$_{0}$&0.222$_{0}$&0.442$_{0}$\\
         \hline
         \textbf{BERT-b} & 0.569$_{21}$& 0.539$_{20}$& 0.770$_{1}$\\
         \hline
         \textbf{BERT-l} & 0.569$_{32}$& 0.558$_{21}$& 0.738$_{32}$\\
         \hline
         \textbf{RoBERTa-b}&0.544$_{14}$&0.527$_{19}$&0.689$_{14}$\\
         \hline
         \textbf{RoBERTa-l}&0.524$_{29}$&0.585$_{17}$&0.754$_{4}$\\
         \hline
         \textbf{Falcon}&0.452$_{11}$&0.567$_{18}$&0.623$_{7}$\\
         \hline
         \textbf{GPT-4}&0.666&\textbf{0.739}&0.852\\
         \hline
         \textbf{Flan-T5-xl}&\textbf{0.684$_{17}$}&0.621$_{17}$&\textbf{0.869$_{16}$}\\
         \hline
         \textbf{Flan-T5-xxl}&0.633$_{16}$&0.655$_{17}$&0.787$_{16}$\\
    \end{tabular}}
    \caption{Indirect scalar intensity ranking table. Subscripted numbers are the best-performing template number. The best results across models are in bold.}
    \label{tab:indirect_intensity}
\end{table}

We found that models all encode rich information about adjective intensity by significantly outperforming the baseline. For BERT and RoBERTa, we again find that direct and indirect probing show similar relative performance, with absolute results for indirect probing lower than those for direct probing.

GPT-4 dominates again in this task as it largely outperforms all models in one dataset and has competitive performance with Flan-T5-xl in the other two. The scaling law does not explain the pattern of results; across different families, Falcon is much larger than BERT and RoBERTa models but performs much worse. Even within encoder models, RoBERTa also fails to outperform BERT.

\section{Probing Scalar Diversity Pragmatics}
\label{sec:scalar-diversity}
Given that LLMs have different levels of lexical-semantic knowledge about SAs, we can now ask whether these different levels correlate with the LLMs' ability to draw pragmatic inferences. This section assesses one particular aspect of this question, namely whether models can reason about the scalar diversity of SAs.

\subsection{Scalar Diversity Probing Models} 
In this section, we describe how we use naturalistic settings to probe LLMs to answer questions about scalar diversity. We only use instruction-tuned models in the previous sections, namely Falcon, GPT-4, Flan-T5-xl/xxl so that they can properly follow instructions.

\subsection{Scalar Diversity Probing Dataset}

We use all SI instances triggered by SAs from \textbf{PVT} \citelanguageresource{PVT}, \textbf{GZ} \citelanguageresource{GZ}, and \textbf{RX} \citelanguageresource{RX}, providing a total of 152 instances. About 40 human participants answer yes or no to each prompt such as 'Mary: The problem is \emph{hard}. Would you conclude from this that Mary thinks the problem is not \emph{unsolvable}?' Since 40 annotators may not represent very fine-grained human judgments, we convert these datasets to binary classifications: we force models to answer \texttt{yes} or \texttt{no} given a question similar to the approach used in \citet{ruis2024goldilocks} (see the next subsection for an illustration), and consider the answers to be \emph{yes} for instances to which at least half of the participants infer SIs, otherwise \emph{no}.

An overview of these datasets can be found in Table \ref{tab:scalardiversity}.

\begin{table}[ht]
    \centering

    \begin{tabular}{c|c|c|c}
         \textbf{Dataset} & \textbf{Total instance} & \textbf{Yes} & \textbf{No}\\
         \hline
         \textbf{PVT} & 50 & 13&37\\
         \hline
         \textbf{GZ} & 70 & 19&51\\
         \hline
         \textbf{RX} & 32 & 5&27\\
    \end{tabular}
    \caption{Overview of total instances and answer counts in scalar diversity datasets.}
    \label{tab:scalardiversity}
\end{table}

The above datasets are relatively small. However, there are no existing large-scale datasets for scalar diversity triggered by SAs. By using three such datasets, we hope to provide a reasonable evaluation of the models' capabilities.

\subsection{Scalar Diversity Probing Method}

We note that some models may have inherent preferences for \emph{yes} or \emph{no} (e.g. more likely to answer \emph{yes} to neutral prompts or vice versa). To debias them for such preferences, we adopt the following strategies. First, we feed the full prompt to models and record the probability for \emph{yes} and \emph{no} as the immediate token following the prompt and denote them as \emph{sy} and \emph{sn}, respectively. For models other than GPT-4, we apply three strategies to retrieve their best performance: (i) \emph{sy}, (ii) weighted probability for \emph{sy} (denoted as \emph{wy}) and (iii) calibrated probability of \emph{wy} (denoted as \emph{cy}). For each strategy, we consider the models' answer to be \emph{yes} if the probability is at least 0.5,  \emph{no} otherwise.

The value of \emph{wy} is calculated as Equation~\ref{eq:wy}:
\begin{equation}
\label{eq:wy}
wy = \frac{sy}{sy+sn}
\end{equation}

\emph{cy} is the calibrated probability of \emph{wy}, which we calibrate with the probability distribution after softmax as \citet{calibrate-before-use} in practice. We use three neutral fillers [N/A], empty string, and [MASK] in positions of scalar claims, which give us six different neutral contexts for each template. An example is provided below, where the black part is prompt and the blue part is the answer option.

\begin{quote}
\texttt{\textbf{Question:}Imagine that your friend Mary says, "[MASK]"}
\\
\texttt{Would you conclude from this that Mary thinks [N/A]?}
\\
\texttt{Only answer yes or no.}
\\
\texttt{\textbf{Answer:} {\color{blue} yes/no}}
\end{quote}

Given these neutral prompts, we average the probabilities of \emph{wy} in each prompt and calculate a calibration weight matrix $W$ which calibrates the averaged \emph{wy} to be 0.5 (i.e., neutralize it). We then use this matrix to calibrate \emph{wy} in non-neutral prompts to derive \emph{cy}.

\subsection{Scalar Diversity Experiment and Results} 
Because the datasets are unbalanced, we use macro F1 as the evaluation metric. We use a logistic regression (LR) model with string-based and concept-based surprisal features \cite{10.1162/tacl_a_00579} as a baseline.\footnote{We drop 4 datapoints in \textbf{GZ} in the baseline as they do not have the surprisal information.} When making SI predictions in one dataset, we use either of the remaining datasets or both of them to train an LR model. We report the best performance and strategies for all models in Table \ref{tab:scalar_diversity_res}.\footnote{See full results for non-baseline models (except GPT-4, which we do not have access to its probability distribution by the time we conduct this experiment) in Appendix \ref{sec:scalar_diversity_res}.}

\begin{table}[ht]
    \centering
\resizebox{\linewidth}{!}{%
    \begin{tabular}{c|c|c|c|c}
        \textbf{Model} & \textbf{RX} & \textbf{GZ} &\textbf{PVT} & \textbf{Avg}\\
        \hline
        \textbf{LR}&0.726$_{RX}$&0.713$_{RX+PVT}$&0.688$_{GZ}$&0.709\\
        \hline
        \textbf{Falcon} & 0.458$_{sy}$& 0.534$_{cy}$& 0.578$_{cy}$&0.464$_{cy}$\\
        \hline
        \textbf{Flan-T5-xl}&0.835$_{cy}$&0.757$_{cy}$&\textbf{0.746$_{cy}$}&0.779$_{cy}$\\
        \hline
        \textbf{Flan-T5-xxl}&\textbf{0.897$_{sy}$}&\textbf{0.867$_{cy}$}&0.726$_{cy}$&\textbf{0.816$_{cy}$}\\
        \hline
        \textbf{GPT-4}&0.759&0.598&0.729&0.695\\
    \end{tabular}}
    \caption{Macro-F1 scores for all models across datasets. Subscripted letters for the baseline model (LR) refer to its training dataset. The average result for LR is taken over the best per-dataset results as a universal training strategy for all datasets does not exist. Subscripted letters for non-baseline models refer to strategies used. The best results across models are in bold.}
    \label{tab:scalar_diversity_res}
\end{table}

Generally, we find that decoder models have unsatisfying performance in this task. For instance, Falcon performs below baseline, and GPT-4 only performs on par with the baseline and underperforms Flan-T5 models. Flan-T5-xxl shows the best results, despite having much space to reach ceiling performance. This result is slightly different from generalized implicature probing results from \citet{ruis2024goldilocks}, which reports that GPT-4 outperforms Flan-T5 models.\footnote{We hypothesize this result could be for two reasons. First, their dataset contains very few direct scalar adjective comparisons. Second, we report zero-shot results instead of few-shot and chain-of-thought results, although \citet{ruis2024goldilocks} notes that these techniques do not help GPT-4 performance.}

\section{A Critical Appraisal of Results}
\subsection{`Bad' Prompts, Good Results}
Recall the following surprising result from \S \ref{sec:scalar-intensity}: using simply concatenated adjective strings as inputs -- which are unnatural -- yields better direct scalar intensity ranking results than natural contexts with single adjectives (G\&A).

From the perspective of human language processing, recent experimental linguistic works show that humans derive more robust scalar contrasts when both strong and weak terms appear salient in the same context \cite{degreeestimateronaixiang}. It seems reasonable to assume that on different days, a speaker may describe a movie as \emph{good} or \emph{awesome}, but actually have the same degree estimate mentally (e.g. \emph{the movie is a 7/10}) despite these terms having different semantic intensities in general. However, when using these contrastive adjectives in the same context, it is less likely that they have the same mental degree estimates. LLMs may work similarly when computing adjective intensities.

\begin{figure}[ht]
    \centering
    \includegraphics[scale=0.4]{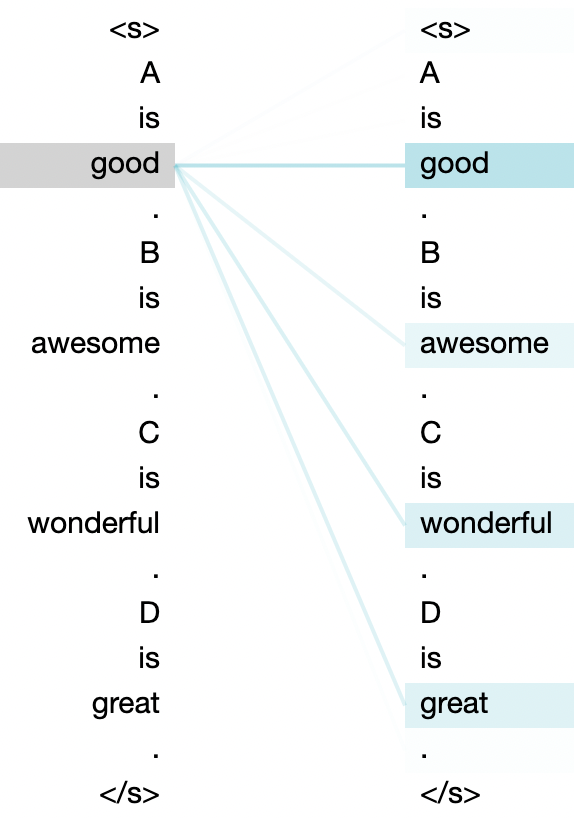}
    \caption{Attention visualization by Bertviz \cite{vig-2019-multiscale}. Attention head 10 in the last layer of RoBERTa-b picks up \emph{good}, \emph{great}, \emph{wonderful}, \emph{awesome} when computing \emph{good} in the context of 'A is good. B is awesome. C is wonderful. D is great.'}
    \label{fig:2}
\end{figure}

However, we also note that previous works suggest that LLMs tend to memorize information from training \cite{zhong-etal-2021-factual} while our prompts are not natural at all (e.g. \emph{good awesome wonderful great} does not exist in Google Ngram corpus). It is unclear why the `bad' prompting method performs better than natural G\&A templates. We hypothesize that attention only picks up salient discourse elements (i.e. the comparative adjectives), which is visualized in Figure \ref{fig:2}. Natural contexts containing these adjectives, which might appear in training, may not differ as much from our prompt when computing SA representations.

\subsection{Indirect Probing: Lower-bounded Absolutely, Faithful Relatively}
Indirect probing shows unsatisfying absolute results compared to direct probing in \S \ref{sec:lexical-semantics}. This finding aligns with \citet{petroni-etal-2019-language}, albeit the relative results remain unchanged. There are several possible explanations. First, although LLMs encode SA lexical semantics in word embeddings, they cannot reason about them in complex linguistic constructions. Second, direct probing in our study utilizes the best layer-wise information. The indirect method may not be as good for comprehensively selecting useful information across model layers. Third, although we have many prompts to get the best performance out of models, they could nevertheless be biased by the training corpus distribution if these prompts are not frequent enough \cite{zhong-etal-2021-factual}.

\subsection{Scaling Does Not Explain Performance}
In the lexical-semantic task (\S \ref{sec:lexical-semantics}), the scaling law tends to hold. However, it does not explain all the patterns in the results: RoBERTa underperforms smaller BERT, Falcon underperforms smaller encoder models, and Flan-T5-xl. \citet{notwacky} also reports similar observations that BERT has better representations of scalar adverbs than RoBERTa. We hypothesize that this can be explained by training objectives. First, models that model sentence relations in pre-training tend to learn fine-grained lexical semantics easier than those without: BERT features next-sentence prediction, and Flan-T5 models sentence relations in the encoders and decoders, which forces them to learn semantics contrastively, which aligns with the findings in \citet{merrill2024can}. Moreover, the fact that Falcon underperforms RoBERTa, which also lacks sentence relation training, can be explained by that the decoders only model left-hand context, which constrains them from reasoning about useful right-hand contexts. However, it is nevertheless surprising that Flan-T5-xxl underperforms the smaller Flan-T5-xl. We leave this question for future investigations.

\subsection{Good Lexical Semantics Does Not Entail Good Pragmatics}

\begin{figure}[ht]
    \centering
    \includegraphics[scale=0.2]{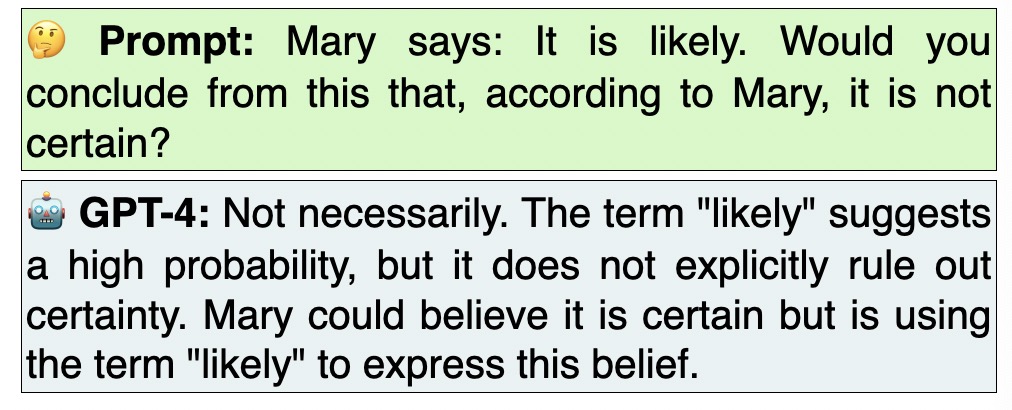}
    \caption{Free generation results for GPT-4 using a prompt from \textbf{GZ} without forcing \texttt{yes} or \texttt{no} answers.}
    \label{fig:free-gen}
\end{figure}

In scalar diversity reasoning (section \ref{sec:scalar-diversity}), GPT-4 performs worse than the much smaller Flan-T5 models, even though it is the best-performing model in the lexical-semantic task. We find that GPT-4 predicts \emph{no} for at least 90\% of the instances. Instead of forcing a \emph{yes} or \emph{no} answer in section \ref{sec:scalar-diversity}, we try free generation here to gain further insights. Figure \ref{fig:free-gen} displays one instance where humans are confident in saying \emph{yes}, but GPT4 says \emph{no}. We see that the model is quite conservative in making statements about implicit status.\footnote{Although conversational implicatures can be canceled \cite{grice1975logic} for discussion), we wish to emphasize that the paper investigates the degree to which LLM behavior matches that of an independently collected, normed, dataset. The possibility of cancellation is less evident to the experimental participants in the dataset collection.}

Considering linguistic capabilities, we may not always expect models that are good at one task to be good at another. \citet{weissweiler-etal-2022-better} also note that better syntax does not always mean better semantics. Specifically, the gap between semantic and pragmatic capabilities in our work can be explained by the fact that a weaker scalar alternative does not always give rise to the negation of the stronger one. Humans consider various factors in addition to lexical semantics such as the discourse context, the frequency of the implicature \cite{pvt}, politeness \cite{gotzner2021face}, and other social factors. Even more confoundingly, humans also need to consider the possibility that conversational implicatures might be canceled in the subsequent discourse \cite{potts2015presupposition}. Due to these complicated concerns, models need to both understand SA lexical semantics and reason pragmatically about various contextual factors to derive implicatures. Thus, good semantic models may not be good pragmatic models in reasoning about communicative intentions.

\section{Conclusion}
This paper has shown that LLMs of different sizes and architectures encode rich information about SA lexical semantics. Moreover, we have shown that good lexical-semantic knowledge of SAs does not always give rise to good performance in the pragmatic reasoning task about their scalar diversity. Finally, we leveraged both linguistic intuitions and model training objectives to provide an analysis for our probing results.

\section{Limitations}
One limitation of our research is that we only use a binary classification for labels. We recognize that scalar diversity can be more fine-grained than simple \emph{yes} or \emph{no}. Moreover, we also note that the datasets that we use for scalar diversity task are relatively small and only constrained to SAs, which is due to the sparsity of large-scale scalar reasoning datasets\footnote{Some datasets containing SI triggers such as determiners and verbs, but they are also small \cite{rx, vt2016}.} We encourage future works to undertake larger-scale collections of fine-grained datasets to evaluate models and train them to suit relevant needs.

Finally, some of our test datasets (e.g. ukWac) are published before the release of our models. LLMs may have already seen the test materials. As discussed in \cite{la2024code, drinkall2024timegpt}, evaluations of LLMs may be artifactually high because of the difficulties in ensuring that the test data are truly invisible during training. Future works can investigate this problem further.

\section{Acknowledgements}
FL is supported by Clarendon and Jason Hu studentships. JBP is supported
by the Engineering and Physical Sciences Research Council (EP/T023333/1). This work is supported by research funding provided by St Catherine's College and the Faculty of Linguistics, Philology, and Phonetics at the University of Oxford. We are grateful to all the people who have offered generous help and feedback on all versions of this paper.

\section{Bibliographical References}\label{sec:reference}

\bibliographystyle{lrec-coling2024-natbib}
\bibliography{lrec-coling2024-example}

\section{Language Resource References}
\label{lr:ref}
\bibliographystylelanguageresource{lrec-coling2024-natbib}
\bibliographylanguageresource{languageresource}
\clearpage

\onecolumn
\section{Appendix}
\subsection{Alternative Methods for Scale Membership Probing}
\label{sec:alternative_membership}
Instead of summing up the weakest and strongest adjectives on each half-scale as the scale vector, and ranking the cosine similarity between each adjective and scale vectors for their scale memberships, we also experimented with computing the cosine similarity between each adjective and all adjectives on different scales, and averaging the similarity (we skip the assessed adjective itself when computing similarity with the scale it belongs to). We report results for the best-performing layers below. We observe that using the end adjectives provides better results than all adjectives (i.e. the method in the main content).

\begin{table}[H]
    \centering
    \begin{tabular}{c|c|c|c}
         \textbf{Models}& \textbf{DM} & \textbf{CD} & \textbf{WK} \\
         \hline
         \textbf{fast-text}& \textbf{0.748}& 0.445& \textbf{0.958}\\
         \hline
         \textbf{BERT-b} &0.660$_{\pm0.009}$&0.461$_{\pm0.020}$&0.941$_{\pm0.012}$\\
         \hline
         \textbf{BERT-l} &0.707$_{\pm0.013}$&\textbf{0.481$_{\pm0.014}$}&0.938$_{\pm0.013}$\\
         \hline
         \textbf{RoBERTa-b}&0.342$_{\pm0.021}$&0.203$_{\pm0.015}$&0.612$_{\pm0.043}$\\
         \hline
         \textbf{RoBERTa-l}&0.535$_{\pm0.016}$&0.326$_{\pm0.016}$&0.856$_{\pm0.019}$\\
    \end{tabular}
    \caption{Direct scale membership probing results by comparing averaged cosine similarities between the target adjective to all adjective members on different scales (in comparison to calculating the cosine similarity between the target adjective and the scale vector in the main content). Subscripted numbers are standard deviation in ten runs. The best results per dataset are in bold.}
    \label{tab:alternative_membership_scale_vector}
\end{table}

Next, we show some alternatives to mean pooling, the pooling method used in the main content. In contextualized representation computation, we used mean pooling when words are segmented. Instead of mean pooling, we also tried max/min pooling. We report the results below (we use end adjectives in this experiment). Generally, mean pooling performs better.

\begin{table}[H]
    \centering
    \begin{tabular}{c|c|c|c}
         \textbf{Models}& \textbf{DM} & \textbf{CD} & \textbf{WK} \\
         \hline
         \textbf{BERT-b} &0.815$_{\pm0.009}$/0.807$_{\pm0.006}$&0.783$_{\pm0.011}$/0.775$_{\pm0.010}$&0.997$_{\pm0.004}$/0.996$_{\pm0.004}$\\
         \hline
         \textbf{BERT-l} &0.835$_{\pm0.011}$/0.832$_{\pm0.011}$&0.787$_{\pm0.010}$/0.789$_{\pm0.009}$&0.998$_{\pm0.003}$/0.998$_{\pm0.003}$\\
         \hline
         \textbf{RoBERTa-b}&0.642$_{\pm0.009}$/0.641$_{\pm0.008}$&0.682$_{\pm0.007}$/0.684$_{\pm0.006}$&0.881$_{\pm0.023}$/0.881$_{\pm0.016}$\\
         \hline
         \textbf{RoBERTa-l}&0.727$_{\pm0.010}$/0.723$_{\pm0.006}$&0.721$_{\pm0.007}$/0.720$_{\pm0.008}$&0.954$_{\pm0.011}$/0.950$_{\pm0.014}$\\
    \end{tabular}
    \caption{Direct scale membership probing results using min/max pooling (left/right). Subscripted numbers are standard deviation in ten runs.}
    \label{tab:alternative_membership_pooling}
\end{table}

\subsection{Full Table of Scalar Intensity Ranking Results}
\label{sec:intensity_ranking_full_res}
\begin{table}[H]
    \centering
\resizebox{\linewidth}{!}{%
    \begin{tabular}{c|c|c|c|c|c|c|c|c|c|c|c}
         Model&&&\multicolumn{3}{c}{DM}&\multicolumn{3}{|c}{CD}&\multicolumn{3}{|c}{WK}  \\
         \hline
         &\textit{Dvec}&Method&P-ACC&$\tau$&$\rho$&P-ACC&$\tau$&$\rho$&P-ACC&$\tau$&$\rho$\\
         \Xhline{5\arrayrulewidth}
         \multirow{4}{*}{BERT-base}&\multirow{2}{*}{DM}&G\&A&-&-&-&\underline{0.735}&\underline{0.668}&\underline{0.745}&0.902&0.803&0.820\\
         &&Ours&-&-&-&0.706&0.604&0.681&\textbf{\underline{0.967}}&\textbf{\underline{0.934}}&\textbf{\underline{0.951}}\\\cline{2-12}
         &\multirow{2}{*}{CD}&G\&A&\underline{0.646}&\underline{0.431}&0.509&-&-&-&0.852&0.705&0.756\\
         &&Ours&0.639&0.417&\underline{0.522}&-&-&-&\underline{0.902}&\underline{0.803}&\underline{0.883}\\\cline{2-12}
         &\multirow{2}{*}{WK}&G\&A&\underline{0.584}&\underline{0.303}&0.317&\underline{0.704}&\underline{0.602}&\underline{0.685}&-&-&-\\
         &&Ours&\underline{0.584}&0.295&\underline{0.360}&0.694&0.581&0.659&-&-&-\\
         \Xhline{5\arrayrulewidth}
         \multirow{2}{*}{BERT-large}&\multirow{2}{*}{DM}&G\&A&-&-&-&\underline{0.731}&\underline{0.663}&\underline{0.717}&\underline{0.918}&\underline{0.836}&0.815\\
         &&Ours&-&-&-&0.727&0.650&0.703&0.902&0.803&\underline{0.873}\\\cline{2-12}
         &\multirow{2}{*}{CD}&G\&A&\textbf{\underline{0.695}}&\textbf{\underline{0.531}}&\textbf{\underline{0.623}}&-&-&-&\underline{0.918}&\underline{0.836}&0.868\\
         &&Ours&0.673&0.488&0.606&-&-&-&\underline{0.918}&\underline{0.836}&\underline{0.900}\\\cline{2-12}
         &\multirow{2}{*}{WK}&G\&A&0.613&0.362&0.372&\underline{0.707}&\underline{0.605}&0.649&-&-&-\\
         &&Ours&\underline{0.628}&\underline{0.391}&\underline{0.445}&0.706&\underline{0.605}&\underline{0.685}&-&-&-\\
         \Xhline{5\arrayrulewidth}
         \multirow{2}{*}{RoBERTa-base}&\multirow{2}{*}{DM}&G\&A&-&-&-&0.645&0.485&0.536&0.820&0.539&0.727\\
         &&Ours&-&-&-&\underline{0.748}&\underline{0.698}&\underline{0.759}&\underline{0.934}&\underline{0.869}&\underline{0.929}\\\cline{2-12}
         &\multirow{2}{*}{CD}&G\&A&0.540&0.207&0.222&-&-&-&0.820&0.640&0.710\\
         &&Ours&\underline{0.648}&\underline{0.431}&\underline{0.558}&-&-&-&\underline{0.918}&\underline{0.836}&\underline{0.893}\\\cline{2-12}
         &\multirow{2}{*}{WK}&G\&A&0.557&0.233&0.253&0.599&0.377&0.450&-&-&-\\
         &&Ours&\underline{0.597}&\underline{0.326}&\underline{0.469}&\underline{0.661}&\underline{0.500}&\underline{0.601}&-&-&-\\
         \Xhline{5\arrayrulewidth}
         \multirow{2}{*}{RoBERTa-large}&\multirow{2}{*}{DM}&G\&A&-&-&-&0.682&0.561&0.620&0.836&0.672&0.807\\
         &&Ours&-&-&-&\textbf{\underline{0.752}}&\textbf{\underline{0.702}}&\textbf{\underline{0.783}}&\underline{0.934}&\underline{0.869}&\underline{0.912}\\\cline{2-12}
         &\multirow{2}{*}{CD}&G\&A&0.595&0.323&0.378&-&-&-&0.820&0.639&0.690\\
         &&Ours&\underline{0.664}&\underline{0.465}&\underline{0.612}&-&-&-&\underline{0.902}&\underline{0.803}&\underline{0.880}\\\cline{2-12}
         &\multirow{2}{*}{WK}&G\&A&0.558&0.241&0.261&0.645&0.478&0.571&-&-&-\\
         &&Ours&\underline{0.642}&\underline{0.417}&\underline{0.544}&\underline{0.685}&\underline{0.558}&\underline{0.673}&-&-&-\\
        \end{tabular}}
    \caption{Full evaluation results of applying our methods to different models with additional metrics Kendall’s $\tau$ and Spearman’s $\rho$. The best results across models and methods are marked in bold. Best results using the same model and \textit{Dvec} resource are underlined.}
    \label{tab:full_res_app}
\end{table}

\subsection{Construction templates and examples for scale membership indirect probing}
\label{sec:membership_template}
\begin{table}[H]
    \centering
    \begin{tabular}{c|c|c|c}
         Index&Template & \makecell{Example for adjs \\on the same scale} & \makecell{Example for adjs \\on different scales}\\
         \hline
         1&\textit{ADJ$_{weak}$} \textbf{or even} \textit{ADJ$_{strong}$} & \makecell{\textit{warm} or even \textit{hot}}&\makecell{\#\textit{warm} or even \textit{tasty}}\\
         \hline
         2&\textit{ADJ$_{weak}$} \textbf{ if not} \textit{ADJ$_{strong}$} & \makecell{\textit{warm}, if not \textit{hot}}&\makecell{\#\textit{warm} if not \textit{tasty}}\\
         \hline
         3&\textit{ADJ$_{weak}$}, \textbf{or even} \textit{ADJ$_{strong}$} & \makecell{\textit{warm}, or even \textit{hot}}&\makecell{\#\textit{warm} or even \textit{tasty}}\\
         \hline
         4&\textit{ADJ$_{weak}$}, \textbf{ if not} \textit{ADJ$_{strong}$} & \makecell{\textit{warm}, if not \textit{hot}}&\makecell{\#\textit{warm}, if not \textit{tasty}}\\
    \end{tabular}
    \caption{Construction templates with which adjectives on the same scale are more likely to appear than those on different scales. \# means pragmatically bad expressions.}
    \label{tab:membership_template}
\end{table}

\subsection{All Templates Used for Indirect Scalar Adjective Intensity Ranking Probing}
\label{sec:intensity_template}
\begin{table}[H]
    \centering
    \begin{tabular}{c|c|c|c}
         Index&weak-strong &Index& strong-weak \\
         \hline
         0&ADJ$_{weak}$ but not ADJ$_{strong}$&22&not ADJ$_{strong}$ just ADJ$_{weak}$\\
         \hline
         1&ADJ$_{weak}$ and almost ADJ$_{strong}$&23&not ADJ$_{strong}$, just ADJ$_{weak}$\\ 
         \hline
         2&ADJ$_{weak}$ and even ADJ$_{strong}$&24&not ADJ$_{strong}$ but just ADJ$_{weak}$\\
         \hline
         3&ADJ$_{weak}$ or even ADJ$_{strong}$&25&not ADJ$_{strong}$, but just ADJ$_{weak}$\\ 
         \hline
         4&ADJ$_{weak}$ although not ADJ$_{strong}$&26&not ADJ$_{strong}$ but still ADJ$_{weak}$\\
         \hline
         5&ADJ$_{weak}$ if not ADJ$_{strong}$&27&not ADJ$_{strong}$, but just ADJ$_{weak}$\\
         \hline
         6&ADJ$_{weak}$ though not ADJ$_{strong}$&28&not ADJ$_{strong}$ still ADJ$_{weak}$\\ 
         \hline
         7&ADJ$_{weak}$ or almost ADJ$_{strong}$&29&not ADJ$_{strong}$, still ADJ$_{weak}$\\ 
         \hline
         8&ADJ$_{weak}$, and even ADJ$_{strong}$&30&not ADJ$_{strong}$ although still ADJ$_{weak}$\\ 
         \hline
         9&ADJ$_{weak}$, or even ADJ$_{strong}$&31&not ADJ$_{strong}$, although still ADJ$_{weak}$\\
         \hline
         10&ADJ$_{weak}$, or almost ADJ$_{strong}$&32&not ADJ$_{strong}$, though still ADJ$_{weak}$\\
         \hline
         11&ADJ$_{weak}$, and almost ADJ$_{strong}$&33&not ADJ$_{strong}$ though still ADJ$_{weak}$\\
         \hline
         12&ADJ$_{weak}$ though not ADJ$_{strong}$&&\\
         \hline
         13&ADJ$_{weak}$, although not ADJ$_{strong}$&&\\
         \hline
         14&ADJ$_{weak}$, but not ADJ$_{strong}$&&\\
         \hline
         15&ADJ$_{weak}$, if not ADJ$_{strong}$&&\\
         \hline
         16&not only ADJ$_{weak}$ but ADJ$_{strong}$&\\
         \hline
         17&not just ADJ$_{weak}$ but ADJ$_{strong}$&\\
         \hline
         18&ADJ$_{weak}$ even ADJ$_{strong}$&\\
         \hline
         19&ADJ$_{weak}$ almost ADJ$_{strong}$&\\
         \hline
         20&ADJ$_{weak}$, even ADJ$_{strong}$&\\
         \hline
         21&ADJ$_{weak}$, almost ADJ$_{strong}$&\\ 
    \end{tabular}
    \caption{All templates used in computing scalar adjective intensity ranking.}
    \label{tab:intensity_templates}
\end{table}

\subsection{Held-out Prompt Engineering Results}
In the main content of the paper, we report indirect probing results using the best-performing template for each model. This may incur some concerns about generalizability. Here, we choose the best-performing prompt in the other two datasets when evaluating one dataset. For GPT-4, we simply use the best template found in other models. Our general conclusion remains unchanged.

\begin{table}[h]
    \centering
    \begin{tabular}{c|c|c|c}
         \textbf{Models} & \textbf{DM} & \textbf{CD} & \textbf{WK} \\
         \hline
         \textbf{Google Ngram}&0.287$_{1}$&0.075$_{1}$&0.368$_{1}$\\
         \hline
         \textbf{BERT-b}&0.425$_{1}$&0.066$_{1}$&0.167$_{1}$\\
         \hline
         \textbf{BERT-l}&0.482$_{3}$&0.102$_{1}$&0.250$_{1}$\\
         \hline
         \textbf{RoBERTa-b}&0.266$_{1}$&0.050$_{1}$&0.114$_{1}$\\
         \hline
         \textbf{RoBERTa-l}&0.463$_{1}$&0.100$_{1}$&0.286$_{1}$\\
         \hline
         \textbf{Falcon}&0.265$_{1}$&0.045$_{1}$&0.167$_{1}$\\
         \hline
         \textbf{GPT-4}&0.540&\textbf{0.273}&\textbf{0.500}\\
         \hline
         \textbf{Flan-T5-xl}&\textbf{0.544$_{3}$}&0.177$_{3}$&\textbf{0.500$_{3}$}\\
         \hline
         \textbf{Flan-T5-xxl}&0.515$_{3}$&0.156$_{3}$&0.364$_{3}$\\
    \end{tabular}
    \caption{Indirect probing results for scale membership with the best prompt found in held-out datasets. The best results per dataset are in bold. Subscripted digits are the best template indexes.}
    \label{tab:indirect_membership_app}
\end{table}

\begin{table}[H]
    \centering
    \begin{tabular}{c|c|c|c}
         \textbf{Models} & \textbf{DM} & \textbf{CD} & \textbf{WK}\\
         \hline
         \textbf{Google ngram}&0.312$_{0}$&0.222$_{0}$&0.442$_{0}$\\
         \hline
         \textbf{BERT-b} &0.546$_{1}$& 0.509$_{1}$& 0.623$_{21}$\\
         \hline
         \textbf{BERT-l} &0.529$_{21}$& 0.518$_{32}$& 0.738$_{32}$ \\
         \hline
         \textbf{RoBERTa-b}&0.544$_{14}$&0.527$_{19}$&0.689$_{14}$\\
         \hline
         \textbf{RoBERTa-l}&0.513$_{4}$&0.555$_{4}$&0.721$_{6}$\\
         \hline
         \textbf{Falcon}&0.451$_{18}$& 0.567$_{18}$& 0.607$_{18}$\\
         \hline
         \textbf{GPT-4}&\textbf{0.666}&\textbf{0.739}&\textbf{0.852}\\
         \hline
         \textbf{Flan-T5-xl}&0.648$_{16}$& 0.612$_{16}$& 0.754$_{17}$\\
         \hline
         \textbf{Flan-T5-xxl}&0.564$_{17}$& 0.582$_{16}$& 0.770$_{17}$\\
    \end{tabular}
    \caption{Indirect scalar intensity ranking table. Subscripted numbers are the best-performing template number. The best results across models are in bold.}
    \label{tab:indirect_intensity_app}
\end{table}

\subsection{Model Implementation Details}
\label{sec:implementation_detail}
Flan-T5 and Falcon models are run with float16 precision. All open-source models are run on V100, A100, and M1 chip. GPT-4 is queried via OpenAI API in 2023. Temperature is set to 0 and top\_p is 1 where applicable.

\subsection{Prompts Used for GPT-4}
\label{sec:gpt4_prompt}
In this section, we provide two prompt examples for GPT-4.

\textbf{Adjective scale alignment} \emph{Do not provide explanations. Give five most likely words following the phrase: ADJ$_{weak}$ or even}

\textbf{Adjective intensity ranking} \emph{Prompt: Do not provide explanations. Which of the following phrases is more natural? Answer none if they are equally unnatural. A. not just ADJ$_{weak}$ but ADJ$_{strong}$ B. not just ADJ$_{strong}$ but ADJ$_{weak}$'.}

In adjective intensity ranking, we randomly shuffle the correct answer index to avoid heuristics. The model is expected to return the first phrase as the answer for adjective pairs with unequal intensities, and none for those with equal intensities.

\subsection{Scalar Diversity Results for All Non-baseline Models (except GPT-4) and Strategies}
\label{sec:scalar_diversity_res}
\begin{table}[H]
    \centering
    \begin{tabular}{c|c|c|c|c}
        \textbf{Strategy}&\textbf{Datasets}&\textbf{Falcon}&\textbf{Flan-T5-xl}&\textbf{Flan-T5-xxl}\\
        \Xhline{5\arrayrulewidth}
         \multirow{4}{*}{sy}&RX&\textbf{0.458}&0.714&\textbf{0.897} \\\cline{2-5}
         &GZ&0.421&0.683&0.850 \\\cline{2-5}
         &PVT&0.425&0.688&0.689 \\\cline{2-5}
         &Average&0.435&0.695&0.812\\
         \Xhline{5\arrayrulewidth}
         \multirow{4}{*}{wy}&RX&0.135&0.714&0.855 \\\cline{2-5}
         &GZ&0.323&0.683&0.850 \\\cline{2-5}
         &PVT&0.206&0.688&\textbf{0.726} \\\cline{2-5}
         &Average&0.221&0.695&0.811\\
         \Xhline{5\arrayrulewidth}
         \multirow{3}{*}{cy}&RX&0.281&\textbf{0.835}&0.855 \\\cline{2-5}
         &GZ&\textbf{0.534}&\textbf{0.757}&\textbf{0.867} \\\cline{2-5}
         &PVT&\textbf{0.578}&\textbf{0.746}&\textbf{0.726} \\\cline{2-5}
         &Average&\textbf{0.464}&\textbf{0.779}&\textbf{0.816}\\
    \end{tabular}
    \caption{Full table of scalar diversity results. The best results per model per dataset are marked in bold.}
    \label{tab:scalarimplicature}
\end{table}
\end{document}